**Title:** An Emergency Medical Services Clinical Audit System driven by Weakly-Supervised Named Entity Recognition from Deep Learning

**Authors:** Wang Han* 1 and Wesley Yeung* 2, 3, 4, Angeline Tung 2, 5, Joey Tay Ai Meng 2, Davin Ryanputera 2, 3, Feng Mengling 1, Shalini Arulanadam 2, 6

*Both authors contributed equally

**Corresponding author:** Feng Mengling

**Affiliations:**

1 Saw Swee Hock School of Public Health, National University of Singapore and National University Health System, Singapore

2 Singapore Civil Defence Force, Singapore

3 National University Hospital, Singapore

4 Laboratory for Computational Physiology, Harvard Science and Technology Division, Massachusetts Institute of Technology, Cambridge, MA, USA

5 Home Team Science & Technology Agency, Singapore

6 Singapore General Hospital, Singapore

**Corresponding Author:** Feng Mengling

**Corresponding Address:** 12 Science Drive 2, #10-01, National University of Singapore, Singapore 117549

**Abstract**

**Introduction**

Clinical performance audits are routinely performed in Emergency Medical Services (EMS) to ensure adherence to treatment protocols, to identify individual areas of weakness for remediation, and to discover systemic deficiencies to guide the development of the training syllabus. At present, these audits are performed by manual chart review which is time-consuming and laborious. We develop an automatic audit system based on both the structured and unstructured ambulance case records and clinical notes with a deep neural network-based named entities recognition model.

**Methods**

The dataset used in this study contained 58,898 ambulance incidents encountered by the Singapore Civil Defence Force from 1st April 2019 to 30th June 2019. We employ a weakly-supervised learning approach for model training. In the first step, we used a rule-based technique to label the entire dataset using fuzzy string for longer entities and exact string matching for shorter entities. We trained a Bidirectional Long Short Term Memory-Conditional Random Fields (BiLSTM-CRF) model and fine-tuned two pretrained Bidirectional Encoder Representations from Transformers (BERT) models. We compared the entity level performance of the models using MUC-5 evaluation metrics defined in SemEval'13 on a 2.5% held out test set. We also evaluated the model complexity and the inference speed of the models. A demonstration of our work and model is published on a public website to support validation and reproducibility.

**Results**

The total data set consists of 3,069,578 words with 39,067 unique words. The training, development, and test data sets contain 2,916,211, 76,702, and 76,666 words respectively. All three models achieve F1 scores of around 0.981 under entity type matching evaluation and around 0.976 under strict evaluation. Our BiLSTM-CRF model is 1~2 orders of magnitude lighter and faster than our BERT-based models.

**Conclusion**

Our weakly-supervised training approach yielded a named entity recognition model that could reliably identify clinical entities from unstructured paramedic free-text reports. Our proposed system may improve the efficiency of clinical performance audits and can also help with EMS database research. Further studies are needed to evaluate its performance on data from other EMS systems.

**Introduction**

The Singapore Civil Defence Force (SCDF) is the national emergency ambulance services provider in Singapore handling more than 160,000 medical conveyances per year. Paramedics in SCDF are trained to respond to medical and trauma emergencies by providing rapid on-scene triage, treatment, and conveyance of casualties. Data collected through the course of the patient encounter are routinely documented into an electronic ambulance case record. These records are akin to electronic health record systems that are routinely used in hospitals, and contain data including the time and location of incidents, patient demographics, medical history, chief complaint, vital signs, investigations, physical findings and treatments administered during prehospital care.

Clinical performance audits are routinely performed to ensure adherence to treatment protocols, to identify individual areas of weakness for remediation, and to discover systemic deficiencies to guide development of training syllabus. These audits have traditionally been performed by senior paramedics through a manual chart review. Firstly, the clinical scenario is identified using the chief complaint and other ancillary information. Next the assessment and treatments steps performed are compared with protocols for each clinical scenario. Steps which were not attempted, performed or documented are identified and collated into an audit report. Lastly, feedback on protocol deviations is given to the paramedic, and relevant supervisors for remedial action. This process is both time consuming and labor intensive, resulting in only a small percentage of records being regularly audited.

Part of the problem lies in the fact that most of the treatment and assessment steps are documented in unstructured free text that are analogous to clinical notes in hospital electronic health records. These free text reports contain useful information about clinical findings, patient diagnosis and treatment rendered during patient encounters. However, the unstructured nature of these text reports prevents easy information extraction. They also suffer from challenges that are common for clinical natural language processing (NLP), which include widespread and inconsistent use of acronyms, misspellings, flexible formatting, atypical grammar and use of jargon.[1] Named entity recognition is a NLP technique that has been successfully used for information extraction in medical texts,[2,3] but its application specific to paramedic text reports is unexplored.

In this study, we developed a system to automate the audit workflow by combining information from both the structured and unstructured data contained in ambulance case records and matching them to an existing set of rules used for clinical performance audit. We use three clinical scenarios that are commonly encountered in prehospital practice as motivating examples: i) acute coronary syndrome, ii) stroke, and the iii) bleeding patient. Next, we combine clinical knowledge from prehospital medicine, advances in NLP techniques, and a prehospital electronic health record database to develop a named entity recognition model that extracts clinical entities from unstructured paramedic free text reports.

**Methods**

**Overall System Design**

The information flow within our system is depicted in Figure 1. First, we determine the clinical scenario based on some eligibility criteria. These eligibility criteria may be present in structured fields such as "Chief complaint" or they might be present in unstructured free text such as "Physical finding". Information contained in structured fields can be accessed by simple boolean statements. Information contained in free text is extracted by a named entity recognition model which we will elaborate on below. Gathering information from both sources, we obtain the scenario type and the paramedic actions performed for a case. For each scenario type, the correct actions are actually protocoled, as shown in Table 1. For example, the use of sublingual nitroglycerin for acute coronary syndrome is contraindicated in patients who have systolic blood pressure less than 90mmHg. The paramedics actions are compared to the protocol actions to check if the protocol actions were performed or attempted when indicated. For the purposes of the clinical performance audit, documentation of an attempted action, such as a failed attempt to insert an intravenous cannula or offering a medication is sufficient to generate a pass.

The results of the comparison are then used to compute summary statistics in the form of an audit report which can contain data aggregated at multiple levels. At a case-level, a report can be generated to indicate which actions were performed for a particular incident. At a provider-level, The results can also be aggregated to calculate the frequency for which indicated actions were performed for all cases attended to by a particular provider. At a system-level, the frequency of indicated actions for various clinical scenarios can be calculated to indicate overall performance.

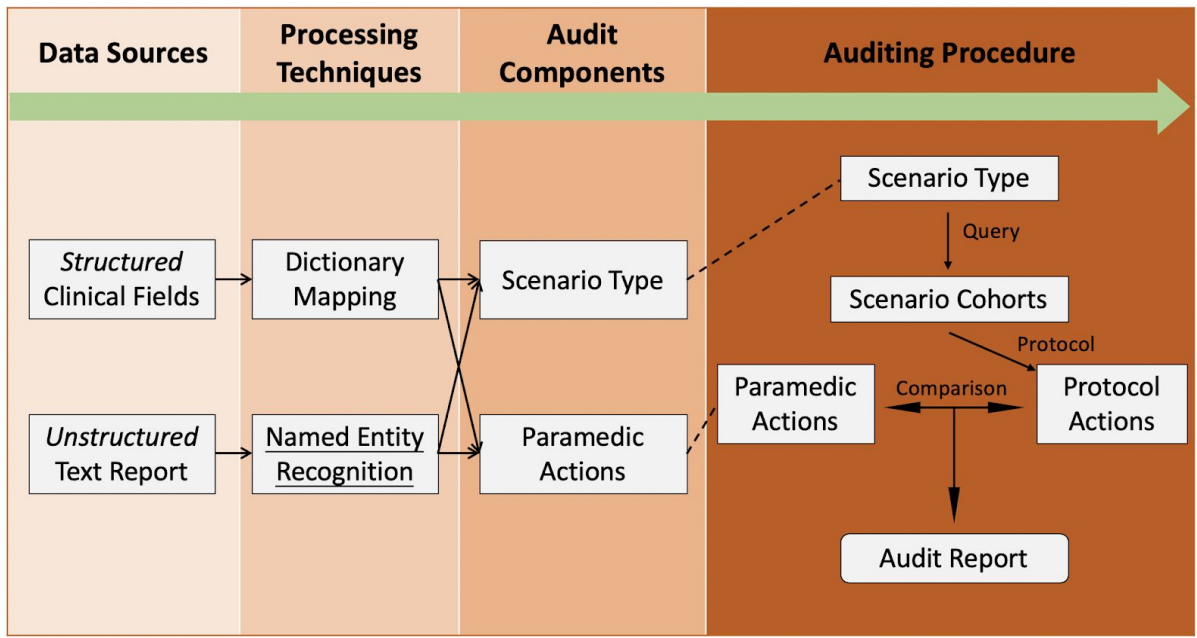

**Figure 1. System architecture and process flow**

| Clinical Category | Scenario Type | Protocol Action |
|---|---|---|
| **Acute Coronary Syndrome**<br><br>Eligibility criteria:<br>- Chief complaint of "Chest Pain" | Default | •Oral aspirin 300mg given<br>•12 lead electrocardiogram performed |
| | + Systolic blood pressure ≥90mmHg | Additional action:<br>•Sublingual nitroglycerin given |
| **Stroke**<br><br>Eligibility criteria:<br>- Chief complaint of "Suspected Stroke" | Default | •Cincinnati prehospital stroke scale performed and recorded<br>•Capillary blood glucose recorded |
| **Bleeding Patient**<br><br>Eligibility criteria:<br>- Physical finding of "Active Bleeding" | Default | •Bleeding control applied (e.g. dressing, manual compression, tourniquet) |
| | + Systolic blood pressure <80mmHg | Additional actions:<br>•IV access established<br>•IV normal saline administered |

**Table 1. Three types of clinical categories, the underlying clinical scenarios for audit and the protocol actions for each scenario. For "acute coronary syndrome" and "bleeding patient", additional actions are required if the systolic blood pressure of the patient is above or below a certain threshold.**

**Named Entity Recognition Module**

In this section, we will introduce how we attempted to build an NER module from unlabelled paramedics text reports.

**Preparation of dataset**

The dataset used in this study contained 58,898 ambulance incidents encountered by the Singapore Civil Defence Force from 1st April 2019 to 30th June 2019. There were 14,679 incidents that did not result in a patient encounter, and 8 cases with missing text reports which were excluded from the study. The remaining 44,211 incidents were split into 95% training (n=42,000), 2.5% development (n=1,105) and 2.5% test (n=1,106) data sets. For our NER labels, we chose 17 different clinical entities spanning 3 different categories (clinical procedure, clinical finding and medication) based on the clinical relevance in EMS practice.

**Text Preprocessing**

Text reports were converted into lower case as many text reports were entered in all upper case due to the nature of the data entry system. All symbols were removed with the exception of the '%' symbol while all numbers were retained. Sentences were split into individual tokens using white space tokenization.

**Weakly-supervised Labelling**

As the entire dataset was unlabelled, we used a weakly-supervised learning approach to model training. In the first step, we used a rule-based technique to label the entire dataset. The notation used in this study is the IOB2 notation which assigns a 'B-' token for the first word of each entity, including terms which comprise of only 1 word; an 'I-' token for each subsequent word within the entity; and an 'O' token for all other tokens not belonging to any entity.[4] We did not annotate negation because we consider the paramedics to be compliant as long as the entity is documented.

We first created a list of synonyms for each entity. These synonyms might be single token or multiple tokens and each entity can have multiple synonyms. Fuzzy string matching was used to increase recall of the bootstrapping process by including terms with minor spelling mistakes. However, as this might result in high rates of false positives if used on short phrases, it was only performed on strings that were 5 characters or longer. The fuzzy string-matching algorithm used was provided by the 'fuzzysearch' package.[5] Single or Multi-token entities that spanned less than 5 characters were matched using exact string matching. If explicitly specified, exact string matching was used.

After labelling of training, development and test sets, each token in development and test sets were verified by a clinician and any mistakes made by the bootstrap process were corrected. We performed error analysis on only the development set to fine-tune the synonym list and specify which terms require exact string matching to improve the labelling process.

**Model Training**

To perform the named entity recognition task, we experimented with a deep learning based Bidirectional Long Short Term Memory + Conditional Random Fields model as well as two Bidirectional Encoder Representations from Transformer (BERT) models with different pretrained weights.

**Bidirectional Long Short Term Memory + Conditional Random Fields**

Conditional Random Fields (CRF) are a class of probabilistic models designed to segment and label sequence data,[6] and have been used with success on named entity recognition tasks due to their ability to use customized observation features from both past and future elements in sequences.[7] Bidirectional Long Short-Term Memory (BiLSTM)[8] combined with an output CRF layer (BiLSTM-CRF)[9] is a recurrent neural network (RNN) model that has achieved state-of-the-art performance over many named entity recognition tasks. Instead of manually crafting features for the traditional CRF model, a BiLSTM model automatically learns the useful features and feeds them into the CRF model. We built a BiLSTM-CRF model using PyTorch library in Python.[10] No pre-trained word embedding was used, instead a word embedding layer was initialized with 0s and trained together with the entire model. Batch size was set as 512. We used Adam optimizer[11] with a default learning rate of 0.001. Early stopping was implemented and would trigger if the validation loss did not decrease in 5 consecutive epochs to prevent overfitting. The maximum training epochs was set as 300. We experimented with different dimensions of the word embedding layer as well as the hidden layer, and used 100 and 64 respectively in the final model which yielded the best performance on the validation set.

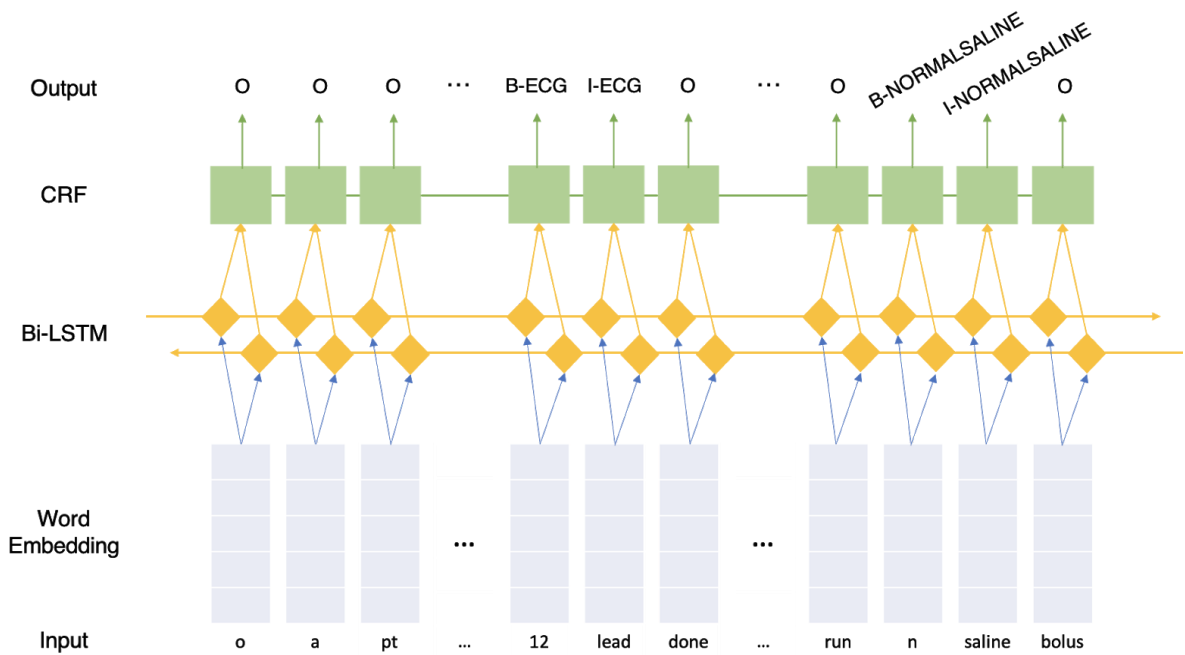

**Figure 2. Illustration of our BiLSTM-CRF model.**

**BERT-based Token Classifier**

BERT-based models are bidirectional transformer models with contextualized word embedding pre-trained on large corpora and have revolutionized deep learning in NLP tasks ever since its introduction.[12] Token classification can be achieved by adding a linear classification layer after the output from the BERT-based model. To build the model, we used the Pytorch implementation from the Transformers Python library by HuggingFace.[13] The first pretrained model we used is the *BERT-base-uncased* model, which was trained on two large corpus BooksCorpus[14] and English Wikipedia. Since our corpus is related to the clinical domain, the second pretrained model we used was Clinical-BERT,[15] which was trained on clinical notes from the Medical Information Mart for Intensive Care III database.[16] Prior to the training, all sentences were tokenized by the pre-trained tokenizer and zero-padded to a constant 300 token sequence length. In the training phase, the pre-trained model was fine-tuned on our training data for 30 epochs with early stopping after 5 epochs if there was no improvement in token level accuracy on the development set. We used AdamW optimizer[17] under the learning rate of $3 * 10^{-5}$, Adams epsilon of $1 * 10^{-8}$ and weight decay rate of 0.01 over all parameters except for the bias terms as well as the gamma and beta terms in the layer-normalization layers. A learning rate scheduler was used to linearly reduce the learning rate throughout the epochs. When the model is combining the sub-token tag predictions, we let the model pick the most frequent class except O to be the final prediction of the word.

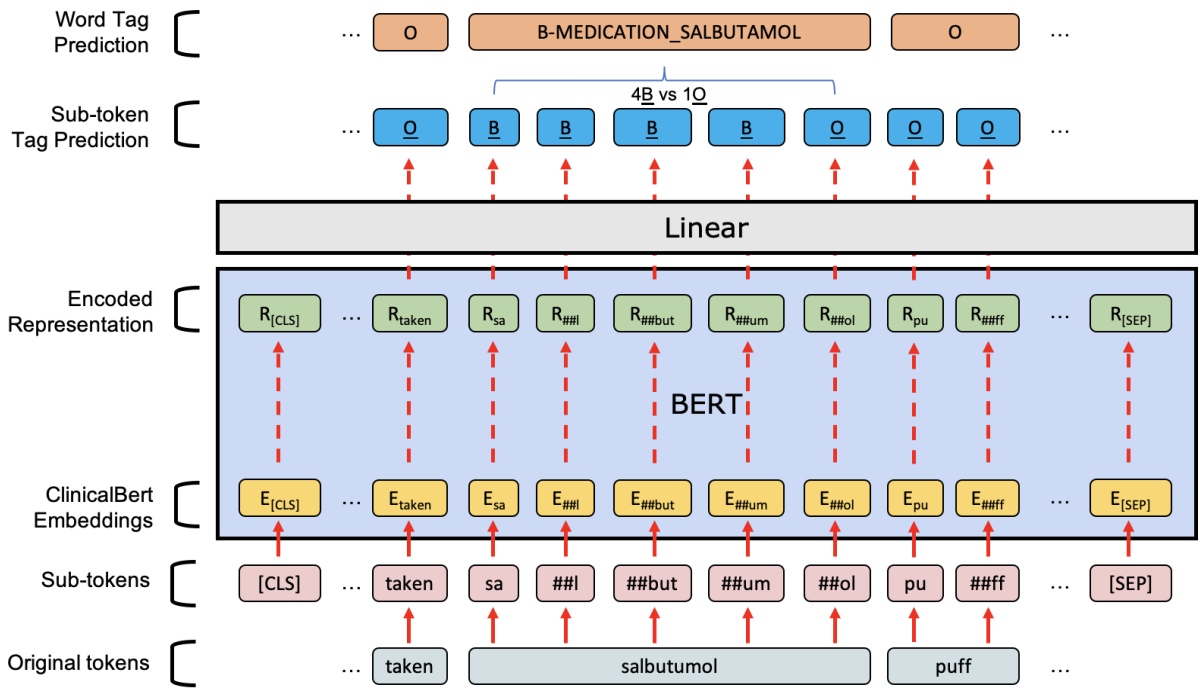

**Figure 3. Illustration of our BERT-based token classification model.**

**Evaluation Metrics**

<u>Token Class Level</u>

We evaluated the performance of our NER models using the weighted precision, recall and F1-score on all tokens except the uninformative 'O' token. Specifically, weighted metric calculates metric for each token class and finds their average weighted by the number of true tokens in that class.

<u>Entity Level</u>

We reported the MUC-5 evaluation metrics under both *strict evaluation* mode and *entity type matching* mode defined in the 2013 International Workshop on Semantic Evaluation (SemEval'13) to compare the performance over the entity level.[18,19] An entity prediction is defined by both the entity type predicted and the word span (starting word and ending word). MUC-5 categorizes each prediction into 1 of the 5 following types:

- Correct (COR) : the system's output is the same as the gold-standard annotation
- Incorrect (INC) : the system's output has nothing in common with the gold-standard annotation
- Partial (PAR) : the system's output shares some overlapping text with the gold-standard annotation
- Missing (MIS) : a gold-standard annotation is not captured by the system
- Spurius (SPU) : the system labels an entity which doesn't exist in the gold-standard annotation

Based on these types, the below measures can be calculated:

- Possible (POS): The number of annotations in the gold-standard which contribute to the final score:
$$POS = COR + INC + PAR + MIS = TP + FN$$
- Actual (ACT): The total number of annotations produced by the system
$$ACT = COR + INC + PAR + SPU = TP + FP$$
- Precision: The percentage of entities found by the system that are correct
$$P = COR / ACT = TP / (TP + FP)$$
- Recall: The percentage of entities present in the data that are found by the system
$$R = COR / POS = TP / (TP + FN)$$

- F1-score: The harmonic mean of precision and recall

$$F1 = 2 * P * R / (P + R)$$

Under both strict evaluation mode and entity type matching mode, there will be no PAR. The only difference between the two modes is that if a predicted entity has the correct type but the word span only overlaps with the gold-standard annotation, it will be INC under strict evaluation, but COR under entity type matching. It is worth noting that POS depends on the model-specific prediction and can be larger than the total number of entities in the data, because 1 gold-standard entity can be compared to more than 1 prediction overlapped with it and be counted more than once.

**Web App Deployment**

We built a publicly accessible website, https://emsnlp.herokuapp.com/, with Flask web application framework,[20] Jinja2 template engine,[21] and Heroku cloud application platform.[22] Users from different emergency medical service providers can try their own paramedics report on our website. Visualization of the predicted entities was done by displaCy.[23]

**Figure 4. Screenshots from our web application. The NER component is introduced in the home page, and visitors can try their own paramedics report in the demo page.**

**Results**

The total data set consists of 3,069,578 words with 39,067 unique words. The training, development and test data sets contain 2,916,211, 76,702 and 76,666 tokens respectively. Table 2 shows some examples of the original reports, the reports after preprocessing and their ground truth NER labels. Statistics about the clinical entities, their relative frequencies, total tokens of the entities and average number of tokens per entity are presented in Table 3.

| Category | Original Report | After preprocessing, ground truth label marked |
|---|---|---|
| Acute coronary syndrome | HX FROM PT C/O CHEST PAIN X 2/7 CRUSHING IN NATURE, NON-RADIATING. NO TRAUMA. NO FALL. O/A PT WAS SITTING, ALERT, CONSCIOUS. PT WAS GTN 1 TAB BY SN. O/E PT NOT PALLOR OR DIAPHORETIC. NO SOB/ GIDDINESS/ NAUSEA/ VOMITTING. AFEBRILE. GIVEN 300MG ASPIRIN STAT DOSE & 1 GTN SPRAY 0.4MG WITH TOTAL RELIEVED. 12 LEAD ECG DONE: SINUS RHYTHM. NO OTHER MEDICAL COMPLAINTS | hx from pt c o chest pain x 2 7 crushing in nature non radiating no trauma no fall o a pt was sitting alert conscious pt was gtn 1 tab by sn o e pt not pallor or diaphoretic no sob giddiness nausea vomitting afebrile given 300mg aspirin stat dose & 1 gtn spray 0 4mg with total relieved 12 lead ecg done sinus rhythm no other medical complaints |
| Stroke | hx from helper, @9m noted pt turns lethargic, but able to enunciate words clearly, @ 12pm, noted pt slurred speech w slight rt facial droop. @2pm, tried to feed pt water, and noted dysphagia, drooling. went to see gp @ 310pm, noted to send to a&e. o/a, pt sitting, gcs 15, slight dementia. no c/o unwell. o/e, noted slight rt facial droop+ slurred speech. no bilateral weakness. pt is off hypertension med for a long time. usual bp @ 115/57 | hx from helper 9m noted pt turns lethargic but able to enunciate words clearly 12pm noted pt slurred speech w slight rt facial droop 2pm tried to feed pt water and noted dysphagia drooling went to see gp 310pm noted to send to a&e o a pt sitting gcs 15 slight dementia no c o unwell o e noted slight rt facial droop slurred speech no bilateral weakness pt is off hypertension med for a long time usual bp 115 57 |
| Bleeding | O/A- PT SITTING CONSCIOUS ALERT. HX FR PT- PT FELL DUE TO SLIPPERY FLOOR, UNSURE HIT WHAT OBJECT NOTED BLEEDING, NO LOC. O/E- NOTED 3CM LACERATION ACTIVE BLEEDING. NOTED DISLOCATED RT SHOULDER, PT CLAIMED NUMBNESS BUT IS DUE TO FALL 2/12 AGO, DID NOT SEE DR. PT UNABLE TO GIVE FURTHUR HX AS HE DOES NOT WISH TO TALK MUCH. | o a pt sitting conscious alert hx fr pt pt fell due to slippery floor unsure hit what object noted bleeding no loc o e noted 3cm laceration active bleeding noted dislocated rt shoulder pt claimed numbness but is due to fall 2 12 ago did not see dr pt unable to give furthur hx as he does not wish to talk much |

**Table 2. Three example text reports under three different case categories before and after preprocessing. The ground truth entities are marked in different colors.**

| Category | Entity | Token Abbreviation | Training | | | | Development | | | | Test | | | |
|---|---|---|---|---|---|---|---|---|---|---|---|---|---|---|
| | | | Entities | % of Total | Tokens | Avg. # of Token per Entity | Entities | % of Total | Tokens | Avg. # of Token per Entity | Entities | % of Total | Tokens | Avg. # of Token per Entity |
| Clinical Procedure | Electrocardiogram | ECG | 26688 | 50.6% | 43661 | 1.64 | 691 | 49.1% | 1160 | 1.68 | 665 | 48.9% | 1101 | 1.66 |
| Clinical Procedure | Stroke Assessment | STROKEASSESSMENT | 6571 | 12.5% | 10102 | 1.54 | 175 | 12.4% | 273 | 1.56 | 200 | 14.7% | 306 | 1.53 |
| Clinical Procedure | Intravenous Cannulation | IVCANNULA | 2054 | 3.9% | 2559 | 1.25 | 64 | 4.5% | 83 | 1.30 | 50 | 3.7% | 69 | 1.38 |
| Clinical Procedure | Burns Cooling | BURNSCOOLING | 57 | 0.1% | 57 | 1.00 | 4 | 0.3% | 4 | 1.00 | 0 | 0.0% | 0 | NA |
| Clinical Procedure | Valsalva Maneuver | VALSALVA | 30 | 0.1% | 44 | 1.47 | 2 | 0.1% | 2 | 1.00 | 0 | 0.0% | 0 | NA |
| Clinical Finding | Bleeding | BLEEDING | 7422 | 14.1% | 8785 | 1.18 | 189 | 13.4% | 230 | 1.22 | 182 | 13.4% | 218 | 1.20 |
| Clinical Finding | Signs Of Obvious Death | OBVIOUSDEATH | 323 | 0.6% | 639 | 1.98 | 10 | 0.7% | 18 | 1.80 | 8 | 0.6% | 16 | 2.00 |
| Medication | Nitroglycerin (GTN) | GTN | 2648 | 5.0% | 3835 | 1.45 | 72 | 5.1% | 104 | 1.44 | 64 | 4.7% | 102 | 1.59 |
| Medication | Aspirin | ASPIRIN | 1644 | 3.1% | 1644 | 1.00 | 51 | 3.6% | 51 | 1.00 | 45 | 3.3% | 45 | 1.00 |
| Medication | Normal Saline | NORMALSALINE | 1371 | 2.6% | 4206 | 3.07 | 41 | 2.9% | 115 | 2.80 | 44 | 3.2% | 135 | 3.07 |
| Medication | Penthrox | PENTHROX | 568 | 1.1% | 568 | 1.00 | 19 | 1.3% | 19 | 1.00 | 14 | 1.0% | 14 | 1.00 |
| Medication | Dextrose/Glucose | DEXTROSE | 447 | 0.8% | 447 | 1.00 | 13 | 0.9% | 13 | 1.00 | 14 | 1.0% | 14 | 1.00 |
| Medication | Adrenaline | ADRENALINE | 412 | 0.8% | 412 | 1.00 | 14 | 1.0% | 14 | 1.00 | 8 | 0.6% | 8 | 1.00 |
| Medication | Diazepam | DIAZEPAM | 394 | 0.7% | 394 | 1.00 | 8 | 0.6% | 8 | 1.00 | 11 | 0.8% | 11 | 1.00 |
| Medication | Salbutamol | SALBUTAMOL | 1794 | 3.4% | 1977 | 1.10 | 50 | 3.6% | 57 | 1.14 | 48 | 3.5% | 55 | 1.15 |
| Medication | Tramadol | TRAMADOL | 310 | 0.6% | 310 | 1.00 | 5 | 0.4% | 5 | 1.00 | 7 | 0.5% | 7 | 1.00 |
| Medication | Syntometrine | SYNTOMETRINE | 45 | 0.1% | 45 | 1.00 | 0 | 0.0% | 0 | NA | 1 | 0.1% | 1 | 1.00 |
| Total (% of Total) | | | 52778 (100%) | | 79685 (2.73%) | | 1408 (100%) | | 2156 (2.81%) | | 1361 (100%) | | 2102 (2.74%) | |

**Table 3. Overview of the named clinical entities in our dataset. Around 2.7% of the tokens are associated with an entity, and the rest are "O" tokens.**

Based on the prevalence of the entities, *Electrocardiogram (ECG)*, *Bleeding* and *Stroke Assessment* are the 3 most commonly observed entities. Looking at the tokens, *Normal Saline* has a significantly longer average number of tokens per entity, because it is often represented by phrases like "i v n s" or "iv ns 0 9%" after the punctuation is removed.

Table 4 shows the performance of our NER models over the entities on the test set. On the test set, our models show indistinguishably excellent performance of F1 scores around 0.981 under entity type matching evaluation and 0.976 under strict evaluation.

| Evaluation Mode | Model | MUC-5 Scoring | | | | | | SemEval'13 Metrics | | |
|---|---|---|---|---|---|---|---|---|---|---|
| | | COR | INC | MIS | SPU | POS | ACT | Precision | Recall | F1-Score |
| Entity Type Matching | BiLSTM-CRF | 1336 | 0 | 25 | 26 | 1361 | 1362 | 0.981 | 0.982 | 0.981 |
| | BERT_BASE | 1343 | 0 | 20 | 31 | 1363 | 1374 | 0.977 | 0.985 | 0.981 |
| | ClinicalBERT | 1343 | 0 | 20 | 30 | 1363 | 1373 | 0.978 | 0.985 | 0.982 |
| Strict Evaluation | BiLSTM-CRF | 1329 | 7 | 25 | 26 | 1361 | 1362 | 0.976 | 0.976 | 0.976 |
| | BERT_BASE | 1335 | 8 | 20 | 31 | 1363 | 1374 | 0.972 | 0.979 | 0.976 |
| | Clinical-BERT | 1334 | 9 | 20 | 30 | 1363 | 1373 | 0.972 | 0.979 | 0.975 |

**Table 4. Entity-level performance of our NER model on the test set. All three models show indistinguishably excellent performance of F1 scores.**

Despite the indistinguishable performance, the model complexity and inference speed differ by 1 ~ 2 order of magnitude between the BiLSTM-CRF model and the BERT-based models, as demonstrated in Table 5. Hence, we decided to choose BiLSTM-CRF as our final model. We reported the performance of the BiLSTM-CRF model over the token classes on the development set and test set in the appendix.

|  | # of model parameters (in millions) | Model checkpoint size (Mb) | Inference wall time (ms) |
|---|---|---|---|
| BiLSTM-CRF | 3.83 | 15 | 40 (7.5) |
| BERT$_{base-uncased}$ | 109.50 | 418 | 274 (16.2) |
| Clinical-BERT | 108.33 | 414 | 286 (25.6) |

**Table 5. Comparison of model complexity and inference speed. Inference wall time is reported using the mean and standard deviation of wall clock time over 100 iterations of predicting a sample sentence, without the overhead of loading libraries and the model itself.**

**Discussion**

Clinical performance audits are a useful tool to evaluate service delivery, and may improve the standard of patient care in EMS systems.[24] We used discrepancies between real work performance by our paramedics and their training syllabus to identify both individual deficiencies as well as systemic deficiencies to tailor individual retraining as well as focus our paramedic education and protocol development. This system allowed us to vastly increase not only the number of cases audited, but also the complexity of the audit as dozens of individual actions could be evaluated. Moreover, this could be achieved in a much shorter period of time and than if a team of human auditors were to perform manual chart reviews. This system allows us to increase the proportion of cases that undergo audit to complete coverage even as we experience increasing demand for EMS services in Singapore while reducing mental fatigue for human auditors.

Another possible use case for this system would be to identify cases for database research. With the digitalization of ambulance data, there is increasing opportunity for large scale data analysis and research. The named entity recognition model was able to identify a limited set of commonly used clinical entities accurately. This information can be used to retrieve cases containing a certain clinical entity. This would both reduce the need for manual chart review which is impractical as the number of cases increase as well as vastly increase the potential sample size for any clinical study. As EMS is an important part of the chain of survival for patients requiring emergency care, there is a need for robust identification of particular case types for downstream research tasks.

The named entity recognition model achieved good performance with a F1 score of 0.981 under entity type matching evaluation and 0.976 under strict evaluation. Although the overall performance was satisfactory, we observed two major sources of errors. The first source is partial capture of the full span of multi-word entities including ECG (e.g. "ecg 4 leads"), GTN (e.g. "s l gtn") and Normal Saline (e.g. "i v n s 0 9%"). Since our training set is labelled only via a weakly-supervised approach while the validation set and test set are labelled by humans, slight discrepancies are expected. Nevertheless, we believe these mistakes are of less significance and will hardly affect the audit result since the entities still get labelled. The second source is due to the misspelled words. A misspelled word can either be non-existent ("salbutumol" vs "salbutamol") or have a different meaning ("facial drop" vs "facial droop"). Our BiLSTM-CRF model will mark the first type of misspelled words as "unknown" where a special word vector will be assigned. As for the second type of misspelled words, they are seen both in their normal context and the misspelled context. As a result, these misspelled words are more difficult to learn for the model and contribute to the errors. With that being said, we also observe that some of these misspelled words are correctly predicted, likely due to the CRF module.

We expected the BERT models can mitigate the issue of misspelling by predicting the entities from the subwords produced by WordPiece tokenization.[25] Moreover, we expected Clinical-BERT would perform better with the pretraining on a clinical corpus. Upon examination of the results, we see that pretrained tokenizers from BERT$_{base}$ and Clinical-BERT did produce different subwords. Based on these subwords, the BERT models predict more entities than the BiLSTM-CRF model. However, both true positive and false positive increase, and we observe higher recall, lower precision and F1 scores similar to the BiLSTM-CRF model. We suspect a few reasons why BERT models did not work better than BiLSTM-CRF in our task. 1) The WordPiece tokenization is not designed to correct spelling errors but rather segmenting the meaningful units from the words. 2) In our paramedics report, the words are highly abbreviated, making the tokenizer less helpful. 3) Clinical notes, which Clinical-BERT was pre-trained on, are still different from our paramedics report to a certain extent.

To our knowledge, this is the first study investigating the use of NLP from paramedic written free text reports. We believe that this work can inspire more NLP applications on novel clinical text. In the meantime, we also recognize some limitations of this work. Firstly, the study was conducted within a single EMS system and further studies are needed to evaluate its external validity. We welcome users from other EMS systems to try their data on our website and feedback the performance. Secondly, we didn't manage to correct the spelling errors in the text reports, which we leave as a future challenge. Thirdly, we didn't experiment with lighter BERT-base models like DistillBERT which are smaller and faster than normal BERT models.[26] Lastly, entity classes such as "Burns Cooling" and "Valsalva Maneuver" were absent in the test data set and could not be evaluated.

Future studies can prospectively evaluate actual deployment of this software in an EMS system in both quantitative and qualitative aspects for both audits and paramedics. Machine learning models used for named entity recognition need to be recalibrated over time to reflect changes in documentation practices of practitioners over time and change in personnel over time. Evaluation of such changes over time could be the focus of future studies. Finally, evaluation of the named entity recognition model on data from other EMS systems will help determine if the performance we observed is generalizable.

**Conclusion**

In this work, we designed an emergency medical services clinical audit system that automates the audit process. Our weakly-supervised training approach on unlabelled dataset yielded a named entity recognition model that could reliably identify clinical entities from unstructured paramedic free text reports. Data extracted from this system may improve the efficiency of clinical performance audits and can also help with EMS database research.

**Acknowledgements**

We thank for the following people for their help:
Ms Doris for helping with the creation of the synonym list.
Mr Gerald, Mr Lix and Mr Benny for helping with manual verification of the development and training set labels.
Mr Qi Yang for extraction of data from the source system.

**References**

1. Leaman R, Khare R, Lu Z. Challenges in clinical natural language processing for automated disorder normalization. *J Biomed Inform*. 2015;57:28-37. doi:10.1016/j.jbi.2015.07.010
2. Bodnari A, Deleger L, Lavergne T, Neveol A, Zweigenbaum P. A Supervised Named-Entity Extraction System for Medical Text. :8.
3. Aramaki E, Miura Y, Tonoike M, Ohkuma T, Mashuichi H, Ohe K. TEXT2TABLE: Medical Text Summarization System Based on Named Entity Recognition and Modality Identification. In: *Proceedings of the BioNLP 2009 Workshop*. Association for Computational Linguistics; 2009:185–192. Accessed May 22, 2020. https://www.aclweb.org/anthology/W09-1324
4. Sang EFTK. Memory-Based Shallow Parsing. *Journal of Machine Learning Research*. 2002;2(Mar):559-594.
5. Einat T. *Fuzzysearch: Fuzzysearch Is Useful for Finding Approximate Subsequence Matches*. Accessed May 22, 2020. https://github.com/taleinat/fuzzysearch
6. Lafferty J, McCallum A, Pereira FCN. Conditional Random Fields: Probabilistic Models for Segmenting and Labeling Sequence Data. :10.
7. McCallum A, Li W. Early results for named entity recognition with conditional random fields, feature induction and web-enhanced lexicons. In: *Proceedings of the Seventh Conference on Natural Language Learning at HLT-NAACL 2003 - Volume 4*. CONLL '03. Association for Computational Linguistics; 2003:188–191. doi:10.3115/1119176.1119206
8. Lample G, Ballesteros M, Subramanian S, Kawakami K, Dyer C. Neural Architectures for Named Entity Recognition. *arXiv:160301360 [cs]*. Published online April 7, 2016. Accessed May 22, 2020. http://arxiv.org/abs/1603.01360
9. Huang Z, Xu W, Yu K. Bidirectional LSTM-CRF Models for Sequence Tagging. *arXiv:150801991 [cs]*. Published online August 9, 2015. Accessed May 22, 2020. http://arxiv.org/abs/1508.01991
10. PyTorch. Accessed May 22, 2020. https://www.pytorch.org
11. Kingma DP, Ba J. Adam: A Method for Stochastic Optimization. *arXiv:14126980 [cs]*. Published online January 29, 2017. Accessed May 22, 2020. http://arxiv.org/abs/1412.6980
12. Devlin J, Chang M-W, Lee K, Toutanova K. BERT: Pre-training of Deep Bidirectional Transformers for Language Understanding. *arXiv:181004805 [cs]*. Published online May 24, 2019. Accessed May 22, 2020. http://arxiv.org/abs/1810.04805
13. *Huggingface/Transformers*. Hugging Face; 2020. Accessed May 22, 2020. https://github.com/huggingface/transformers
14. Zhu Y, Kiros R, Zemel R, et al. Aligning Books and Movies: Towards Story-Like Visual Explanations by Watching Movies and Reading Books. In: ; 2015:19-27. Accessed May 22, 2020. https://www.cv-foundation.org/openaccess/content_iccv_2015/html/Zhu_Aligning_Books_and_ICCV_2015_paper.html
15. Alsentzer E. *EmilyAlsentzer/ClinicalBERT*.; 2020. Accessed May 22, 2020. https://github.com/EmilyAlsentzer/clinicalBERT
16. Johnson AEW, Pollard TJ, Shen L, et al. MIMIC-III, a freely accessible critical care database. *Sci Data*. 2016;3(1):160035. doi:10.1038/sdata.2016.35
17. Loshchilov I, Hutter F. Decoupled Weight Decay Regularization. *arXiv:171105101 [cs, math]*. Published online January 4, 2019. Accessed May 22, 2020. http://arxiv.org/abs/1711.05101
18. Chinchor N, Sundheim B. MUC-5 Evaluation Metrics. In: *Fifth Message Understanding*

*Conference (MUC-5): Proceedings of a Conference Held in Baltimore, Maryland, August 25-27, 1993*. ; 1993. Accessed May 22, 2020. https://www.aclweb.org/anthology/M93-1007

19. Segura-Bedmar I, Martinez P, Zazo MH. SemEval-2013 Task 9 : Extraction of Drug-Drug Interactions from Biomedical Texts (DDIExtraction 2013). :10.
20. Welcome to Flask — Flask Documentation (1.1.x). Accessed May 22, 2020. https://flask.palletsprojects.com/en/1.1.x/
21. Jinja — Jinja Documentation (2.11.x). Accessed May 22, 2020. https://jinja.palletsprojects.com/en/2.11.x/
22. Cloud Application Platform | Heroku. Accessed May 22, 2020. https://www.heroku.com/
23. displaCy · spaCy Universe. displaCy. Accessed May 22, 2020. https://spacy.io/universe/project/displacy
24. Munk M-D, White SD, Perry ML, Platt TE, Hardan MS, Stoy WA. Physician Medical Direction andClinical Performance at an Established Emergency Medical Services System. *Prehospital Emergency Care*. 2009;13(2):185-192. doi:10.1080/10903120802706120
25. Wu Y, Schuster M, Chen Z, et al. Google's Neural Machine Translation System: Bridging the Gap between Human and Machine Translation. *arXiv:160908144 [cs]*. Published online October 8, 2016. Accessed May 22, 2020. http://arxiv.org/abs/1609.08144
26. Sanh V, Debut L, Chaumond J, Wolf T. DistilBERT, a distilled version of BERT: smaller, faster, cheaper and lighter. *arXiv:191001108 [cs]*. Published online February 29, 2020. Accessed May 24, 2020. http://arxiv.org/abs/1910.01108

**Appendix**

Table 6 shows the performance of the BiLSTM-CRF model over the token classes on the development set and test set. It achieves a weighted average F1 score of 0.963 on the development set and 0.980 on the test set.

| Token | Developmernt Set | | | | Test Set | | | |
| --- | --- | --- | --- | --- | --- | --- | --- | --- |
| | Precision | Recall | F1-Score | Support | Precision | Recall | F1-Score | Support |
| B-FINDING_BLEEDING | 0.930 | 0.979 | 0.954 | 189 | 0.978 | 0.989 | 0.984 | 182 |
| B-FINDING_OBVIOUSDEATH | 1.000 | 0.800 | 0.889 | 10 | 1.000 | 0.875 | 0.933 | 8 |
| B-MEDICATION_ADRENALINE | 1.000 | 0.857 | 0.923 | 14 | 1.000 | 1.000 | 1.000 | 8 |
| B-MEDICATION_ASPIRIN | 1.000 | 1.000 | 1.000 | 51 | 1.000 | 0.978 | 0.989 | 45 |
| B-MEDICATION_DEXTROSE | 1.000 | 1.000 | 1.000 | 13 | 1.000 | 1.000 | 1.000 | 14 |
| B-MEDICATION_DIAZEPAM | 1.000 | 1.000 | 1.000 | 8 | 1.000 | 1.000 | 1.000 | 11 |
| B-MEDICATION_GTN | 1.000 | 1.000 | 1.000 | 72 | 0.955 | 1.000 | 0.977 | 64 |
| B-MEDICATION_NORMALSALINE | 0.895 | 0.829 | 0.861 | 41 | 0.930 | 0.909 | 0.920 | 44 |
| B-MEDICATION_PENTHROX | 1.000 | 0.947 | 0.973 | 19 | 1.000 | 1.000 | 1.000 | 14 |
| B-MEDICATION_SALBUTAMOL | 1.000 | 1.000 | 1.000 | 50 | 0.978 | 0.938 | 0.957 | 48 |
| B-MEDICATION_SYNTOMETRINE | 0.000 | 0.000 | 0.000 | 0 | 1.000 | 1.000 | 1.000 | 1 |
| B-MEDICATION_TRAMADOL | 1.000 | 1.000 | 1.000 | 5 | 1.000 | 1.000 | 1.000 | 7 |
| B-PROCEDURE_BURNSCOOLING | 1.000 | 1.000 | 1.000 | 4 | 0.000 | 0.000 | 0.000 | 0 |
| B-PROCEDURE_ECG | 0.956 | 0.981 | 0.969 | 691 | 0.986 | 0.986 | 0.986 | 665 |
| B-PROCEDURE_IVCANNULA | 0.953 | 0.953 | 0.953 | 64 | 0.980 | 1.000 | 0.990 | 50 |
| B-PROCEDURE_STROKEASSESSMENT | 0.934 | 0.977 | 0.955 | 175 | 0.975 | 0.975 | 0.975 | 200 |
| B-PROCEDURE_VALSALVA | 1.000 | 0.500 | 0.667 | 2 | 0.000 | 0.000 | 0.000 | 0 |
| I-FINDING_BLEEDING | 0.975 | 0.951 | 0.963 | 41 | 1.000 | 0.972 | 0.986 | 36 |
| I-FINDING_OBVIOUSDEATH | 1.000 | 0.875 | 0.933 | 8 | 1.000 | 0.875 | 0.933 | 8 |
| I-MEDICATION_GTN | 1.000 | 1.000 | 1.000 | 32 | 0.946 | 0.921 | 0.933 | 38 |
| I-MEDICATION_NORMALSALINE | 0.931 | 0.905 | 0.918 | 74 | 0.978 | 0.956 | 0.967 | 91 |
| I-MEDICATION_SALBUTAMOL | 1.000 | 0.857 | 0.923 | 7 | 0.857 | 0.857 | 0.857 | 7 |
| I-PROCEDURE_ECG | 0.998 | 0.947 | 0.972 | 469 | 0.998 | 0.972 | 0.985 | 436 |
| I-PROCEDURE_IVCANNULA | 0.900 | 0.947 | 0.923 | 19 | 1.000 | 0.947 | 0.973 | 19 |
| I-PROCEDURE_STROKEASSESSMENT | 0.931 | 0.959 | 0.945 | 98 | 1.000 | 0.962 | 0.981 | 106 |
| I-PROCEDURE_VALSALVA | 0.000 | 0.000 | 0.000 | 0 | 0.000 | 0.000 | 0.000 | 0 |
| **Weighted Average** | 0.964 | 0.964 | 0.963 | | 0.985 | 0.975 | 0.980 | |

**Table 6. Token-level performance of our NER model on the development and test set.**